\def\year{2017} %File: formatting-instruction.tex
\documentclass[letterpaper]{article} 
\usepackage{aaai17} 
\usepackage{times}
\usepackage{helvet} 
\usepackage{courier} 
\usepackage{hyperref}       % hyperlinks
\usepackage{url}            % simple URL typesetting
\usepackage{booktabs}       % professional-quality tables
\usepackage{amsfonts}       % blackboard math symbols
\usepackage{nicefrac}       % compact symbols for 1/2, etc.
\usepackage{microtype}      % microtypography
\usepackage{amsmath}
\usepackage{wrapfig,booktabs}
\usepackage{color}
\usepackage{xcolor}
\usepackage{graphicx}
\usepackage{subfigure}
\usepackage{pdfpages}
\usepackage[font=small]{caption}
\title{Multi-Path Feedback Recurrent Neural Networks for Scene Parsing}

\author{Xiaojie Jin$^1$ Yunpeng Chen$^2$ Zequn Jie$^2$　　Jiashi Feng$^2$ Shuicheng Yan$^{3,2}$\\
\small $^1$NUS Graduate School for Integrative Science and Engineering, NUS\\
\small$^2$Department of ECE, NUS \qquad $^3$360 AI Institute\\
\tt\small \{xiaojie.jin, chenyunpeng, jiezequn, elefjia\}@u.nus.edu, yanshuicheng@360.cn}

\begin{document}

\maketitle

\begin{abstract}
	In this paper, we consider the scene  parsing problem and propose a novel \textbf{M}ulti-\textbf{P}ath \textbf{F}eedback recurrent neural network (MPF-RNN) for parsing scene images. MPF-RNN can  enhance the
  capability  of RNNs in modeling long-range context information at multiple levels and  better distinguish pixels that are easy to confuse.  Different from feedforward CNNs and RNNs with only single feedback, MPF-RNN propagates the contextual features learned at top layer through \textit{multiple} weighted recurrent connections to learn bottom features. For better training MPF-RNN, we propose a new  strategy that considers accumulative loss  at multiple recurrent
  steps  to improve performance of the MPF-RNN on parsing  small objects. With these two novel components, MPF-RNN has achieved significant improvement over strong baselines (VGG16 and Res101) on five challenging scene parsing benchmarks, including traditional
    SiftFlow, Barcelona, CamVid, Stanford Background as well as the recently released large-scale ADE20K.
\end{abstract}

\section{Introduction}
\label{introduction}
Scene parsing has drawn increasing research interestdue to its wide applications in many attractive areas like
autonomous vehicles, robot navigation and virtual reality. However, it remains
a challenging problem since it requires solving segmentation,
classification and detection simultaneously. 

\begin{figure}[!t] \centering
\centering
	\includegraphics[width=\linewidth]{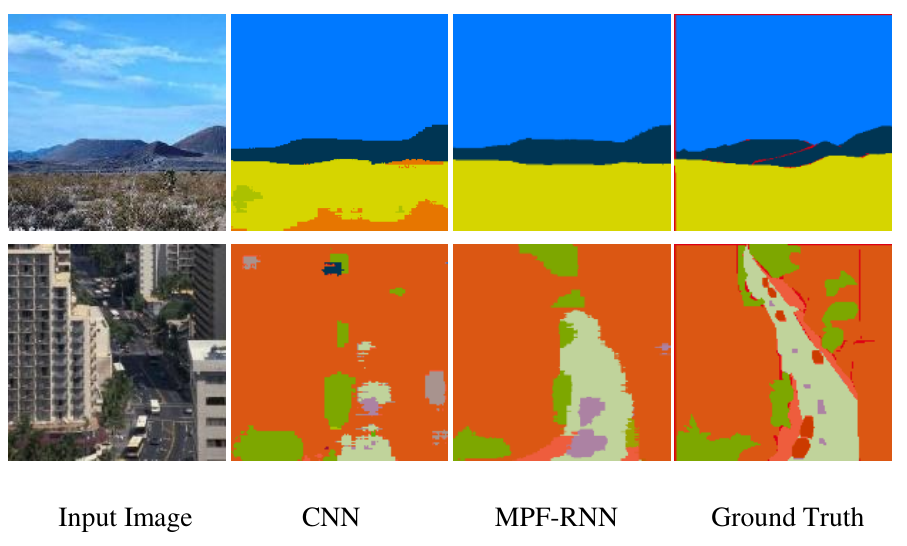}
	\caption{Illustration on importance of context information for distinguishing pixels. Without context, the \textit{field} pixels in the first image are misclassified as \textit{desert}. Similarly, the \textit{street} pixels in the second image are misclassified as \textit{building}.  MPF-RNN can better model context information and thus produces more smooth and semantically meaningful parsing results. Best viewed in color.}
	\label{fig:compare}
\end{figure}

Recently, convolutional neural networks (CNNs) have been widely
used for learning image representations and applied for scene parsing. However,   CNNs  can only capture high-level context information learned at top layers that have large receptive fields (RFs). The bottom layers are not exposed to  valuable context information when they learn features. In addition, several recent works have demonstrated
that even the top layers in a very deep model \emph{e.g.} VGG16~\cite{vgg} have limited  RFs  and receive limited context information  \cite{parsetnet} in fact. Therefore, CNNs usually encounter difficulties in distinguishing  pixels that are easy to confuse locally and
high-level context information is necessary. For example,
in Figure \ref{fig:compare}, without the information of global context, the ``field''
pixels and ``building'' pixels are  categorized incorrectly.

To address this challenge, we propose a novel multi-path feedback recurrent neural
network (MPF-RNN). The overall architecture of our proposed MPF-RNN
is illustrated in Figure \ref{fig:frame}. It has following two appealing characteristics. Firstly, MPF-RNN establishes recurrent
connections to propagate downwards the outputs of the top layer to \emph{multiple} bottom layers
so as to fully exploit the context in the training process of different layers. Benefited from the enhanced context modeling capability, MPF-RNN gains a strong discriminative capability. Note that compared with previous RNN-based
methods \cite{pinheiro2013recurrent,RCNN} which use only one feedback connection from the output layer
to the input layer and the layer-wise self-feedback connections, respectively, MPF-RNN is superior by using a better architecture, \textit{i.e.} explicitly incorporating context information into the training process of multiple
hidden layers, which learns concepts with different abstractness. Secondly,
MPF-RNN effectively fuses the output features across different
time steps for classification or a more concrete parsing purpose. We empirically
demonstrate that such multi-step fusion greatly boosts the final performance of MPF-RNN.

To verify the effectiveness of MPF-RNN, we have conducted extensive experiments
over five popular and challenging scene parsing datasets, including SiftFlow~\cite{siftflow}, Barcelona~\cite{tighe2010superparsing}, CamVid~\cite{brostow2008segmentation}, Stanford Background~\cite{gould2009decomposing} and recently released large-scale ADE20K~\cite{ade20k} and demonstrated that MPF-RNN is capable of greatly
enhancing the discriminative power of per-pixel feature representations.

\section{Related Work}
\label{related_works}

%in the following two aspects.
\paragraph{Image Context Modeling}
One type of context
modeling approaches for scene parsing
is to use the probability graphical models (PGM) (\emph{e.g.} CRF) to improve parsing results. In~\cite{chen2014semantic,zhang2012efficient,roy2014scene,zheng2015conditional,schwing2015fully}, CNN features are combined with a
fully connected CRF to get more accurate parsing results.
Compared to MPF-RNN, such methods usually suffer from intense calculation in inference due to their used time consuming mean field inference, which hinders their application in real-time scenario.
Farabet \emph{et al.}~\cite{farabet2013learning} encoded context information through
surrounding contextual windows from multi-scale images. \cite{socher2011parsing} proposed a recursive neural network to learn a
mapping from visual features to the semantic space for pixel classification
In~\cite{sharma2014recursive}, a parsing tree was used to propagate global context information. In~\cite{parsetnet}, the global context feature is obtained by pooling the last layer's output via global average pooling and then concatenated with local feature maps to learn a per-pixel classifier. Our method is
different from them in both the context modeling scheme and the network architecture and achieves better performance.
%\vspace{-4mm}
\paragraph{Recurrent Neural Networks} RNN has been employed to model long-range context in images. For instance, \cite{pinheiro2013recurrent} built one recurrent connection from the output  to the input
layer, and \cite{RCNN} introduced layer-wise self-recurrent connections.
Compared with those methods, the proposed MPF-RNN models the context  by allowing multiple forms of recurrent connections. %, \emph{i.e.} multiple recurrent connections (including self-recurrent connections) in MPF-RNN.
In addition, MPF-RNN combines the output features at multiple time steps for pixel classification. \cite{stollenga2015parallel} utilized a parallel multi-dimensional long short-term memory for fast volumetric segmentation. However, its performance was relatively inferior. \cite{dag,reseg} were based on similar motivations that used RNNs to refine the learned features from a CNN by modeling the contextual dependencies along multiple spatial directions. In comparison,  MPF-RNN  incorporates the context information into the feature learning process of CNNs.

\section{The MPF-RNN Model}
\label{sec:method}
%We proceed to introduce the proposed MPF-RNN in details.
%Firstly, we give the notations used in the following text and describe the multi-path feedback connections in our model. Secondly, we describe the strategy of
%combining output features at different time steps. Finally, we analyze MPF-RNN by comparing it with previous RNN models to give a better understanding of the proposed method.

\subsection{Multi-Path Feedback}
Throughout the paper, we use the following notations. For conciseness, we only consider one training sample. We denote a training sample as
$(\mathbf{I}, \mathbf{Y})$ where $\mathbf{I}$ is the raw image and $\mathbf{Y}$ is the ground truth parsing map of the image with
$\mathbf{Y}_{i,j} \in \{1,\ldots,K\}$ as the ground truth category  at the
location $(i,j)$ in which $K$ is the number of categories. Since MPF-RNN is built upon CNNs by constructing recurrent connections from the top layer to multiple hidden layers, we therefore firstly consider a CNN composed
of $L$ layers. Each layer outputs a feature map and we denote it as ${\mathbf X}^{\ell}$.
Here ${\mathbf X}^{0}$ and ${\mathbf X}^{L}$ represent the input and final
output of CNN. ${\mathbf W}^{\ell}$ denotes the parameter of filters or weights to
be learned in the $l$-th layer. Using above notations, the outputs of an $L$-layered CNN at each layer can be written as
\begin{small}
\begin{equation}
  \label{activation}
  \footnotesize
  \mathbf{X}^{(\ell)} = f^{(\ell)}(\mathbf{W}^{(\ell)} \mathbf{X}^{(\ell - 1)} ),\ \ \ell =
  1, \ldots ,L\, \ \ \text{and} \ \ \mathbf{X}^{(0)}  \buildrel \Delta \over =  \mathbf{I},
\end{equation}
\end{small}
\protect\noindent where ${\mathbf W}^{(\ell)} \mathbf{X}^{(\ell - 1)}$ performs linear transformations on $\mathbf{X}^{(\ell - 1)}$ and $f^{(\ell)}( \cdot )$ is a composite of multiple specific functions
including the activation function, pooling and
softmax. Here the bias term is absorbed into
$\mathbf{W}^{(\ell)}$.

\begin{figure}
	\centering     %%% not \center
	\subfigure[MPF-RNN in recurrent format]{\label{fig:frame1}\includegraphics[width=\linewidth]{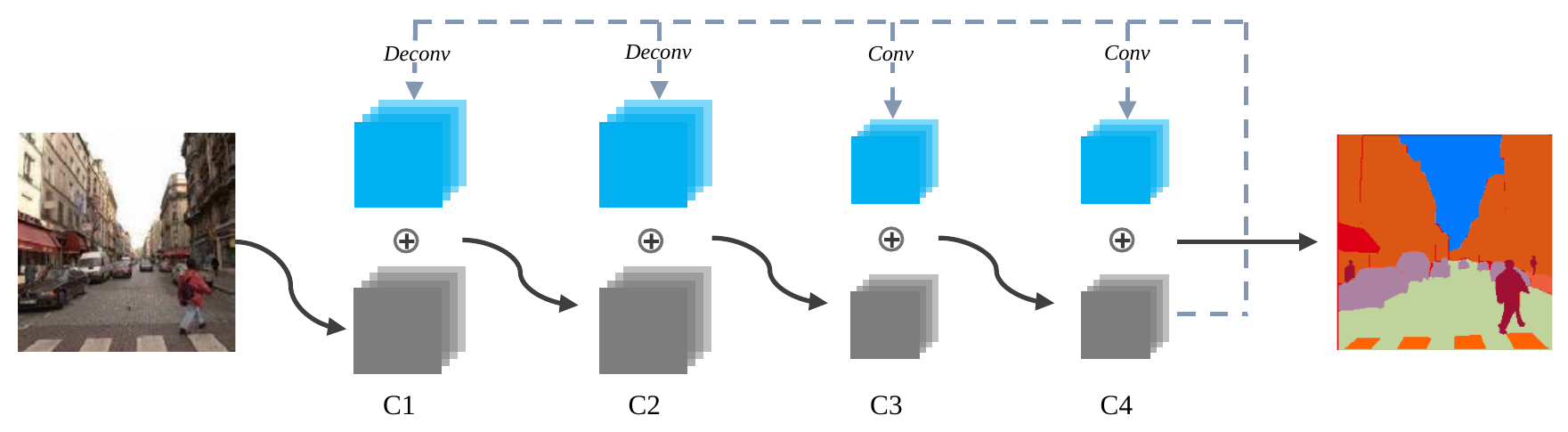}}
	\subfigure[Unfolded MPF-RNN]{\label{fig:frame2}\includegraphics[width=\linewidth]{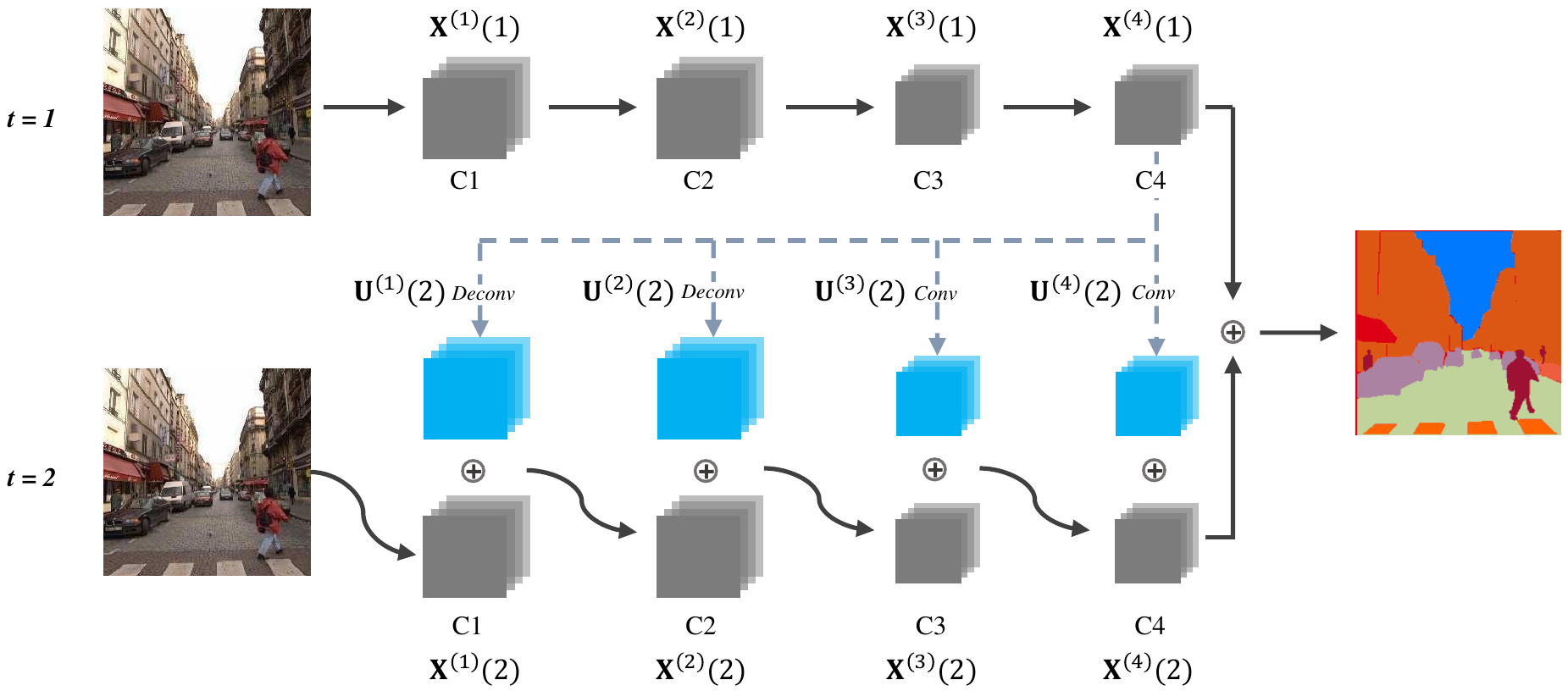}}
	\caption{The framework of MPF-RNN. \textbf{Top}: an MPF-RNN built upon a shallow CNN  of four convolution layers (\emph{i.e.}, C1 to C4) and shown in recurrent format. Multiple feedback connections are constructed from the top layer (C4) to hidden layers, the outputs of which are combined with the back-propagated features via an element-wise sum operation (denoted as $\oplus$). Convolution  and deconvolution layers (\emph{i.e.}, \textit{Conv} and \textit{Deconv}) are used to model the feedback connections from C4 to hidden layers whose outputs have  equal and larger spatial size than C4 respectively. \textbf{Bottom}: the corresponding unfolded MPF-RNN for two time steps. The parameters of convolution layers at different time steps are shared. To be discriminative to small objects, MPF-RNN combines output features across different time steps as the input for the pixel classifier which uses a deconvolution layer to produce the full-size labeling map. All notations in this figure are defined in Eqn. \eqref{eq:mpf} and Eqn. \eqref{eq:loss}.}
	\label{fig:frame}
		%\vspace{-4mm}
\end{figure}

MPF-RNN chooses $M$ layers out of $L$ layers  and constructs
recurrent connections from the top layer to each selected layer.
Let $S = \{ r_m ,m = 1, \ldots ,M\}$ denote the set of selected layers and let $r_m \in \{1, \ldots, L\}$ index the layers. By introducing the recurrent connections, each layer in $S$ takes both the output of its previous layer and the output of the top layer at the last time step as inputs. With $t$ denoting the index of time steps, Eqn. \eqref{activation} can be rewritten as
\begin{small}
\begin{equation}
 % \footnotesize
  \begin{aligned}
    \label{eq:mpf}
    \mathbf{X}^{(\ell)}(t) =
    \begin{cases}
      f^{(\ell)} (g^{(\ell)}(\mathbf{W}^{(\ell)} \mathbf{X}^{(\ell - 1)}(t) )\\  \qquad +
      g^{(\ell)}(\mathbf{U}^{(\ell)}(t) \mathbf{X}^{(L)}(t-1)) ), & \ell \in S, \\
      f^{(\ell)}(\mathbf{W}^{(\ell)} \mathbf{X}^{(\ell - 1)}(t) ), & \text{otherwise},
    \end{cases}
  \end{aligned}
\end{equation}
\end{small}
\\where $\mathbf{X}^{(\ell)}(t)$ and $\mathbf{U}^{(\ell)}(t)$ denote the
output of the $\ell$-th layer and the transformation matrix from the output of the $L$-th
layer to the hidden layer $\ell \in S$ at time step $t$, respectively. Note that
$\mathbf{W}^{(\ell)}$ is time-invariant, which means parameters in weight layers are
shared across different time steps to reduce the model's memory consumption. $\mathbf{U}^{(\ell)}(t)$ is
time-variant so as to learn time step specific transformations for absorbing the context information from top layers. The output of top layers at different time steps conveys context information at different scales thus should have different transformations. The advantages of such a choice  are verified in our experiments.

Following~\cite{parsetnet}, function
$g^{(\ell)}(\cdot)$  normalizes the input
$\mathbf{x}$ as $g^{(\ell)} (\mathbf{x}) = \gamma ^{(\ell)}
\mathbf{x}/||\mathbf{x}||_2$ where $\gamma ^{(\ell)}$ is a learnable scaler and
$||\mathbf{x}||_2$ denotes the $L_2$ norm of $\mathbf{x}$. As verified in \cite{parsetnet}, normalizing two feature maps output at different layers before
performing combination is beneficial for convergence during training. One reason is that those two
feature maps generally have different magnitude scales and they may slow down the
convergence when their magnitudes are not balanced.

Now we proceed to explain the intuitions behind constructing multiple feedback connections. Since the RF of each layer in a deep neural network
increases along with the depth, bottom and middle layers have
relatively small RF. This limits the amount of context information that is perceptible
when learning features in lower layers. Although higher layers might have larger RF and
encode longer-range context, the context captured by higher layers cannot explicitly
influence the states of the units in subsequent layers without top-down
connections. Moreover, according to recent works~\cite{parsetnet,zhoubolei},
the effective RF might be much smaller than its theoretical value. For
example, in a VGG16 model, although the theoretical RF size for the top layer fc7 is equal to $224\times 224$, the effective RF size for the top layer fc7 is only about 1/4 of the theoretical RF. Due to the inadequate context information,
layers in a deep model might not be able to learn context-aware features with strong discriminative power to distinguish local confusing
pixels. Figure \ref{fig:compare}
shows examples of local confusion labeling by a CNN. To make layers in CNN incorporate context information, we therefore propose to propagate the output of the top layer to multiple hidden layers. Modulated with context, the deep model is context-aware when hierarchically
learning concepts with different abstract levels. As a result, the model captures context dependencies to distinguish confusing local pixels. Actually, as the time step increases, the states of every unit in
recurrent layers would have increasingly larger RF sizes and incorporate richer context.%in an arbitrarily large region.

We would also like to explain the reasons why we only build recurrent connections from the last layer: First, the last layer has the largest receptive field among all the layers whose output feature contains the richest contexts. Thus putting other layers in recurrent connections will introduce redundant context information and may hurt performance. Secondly, including more layers significantly increases computation cost. Only applying on the last layer gives a good trade-off between performance and efficiency.

\begin{figure}[!t] \centering
	\includegraphics[width=\linewidth]{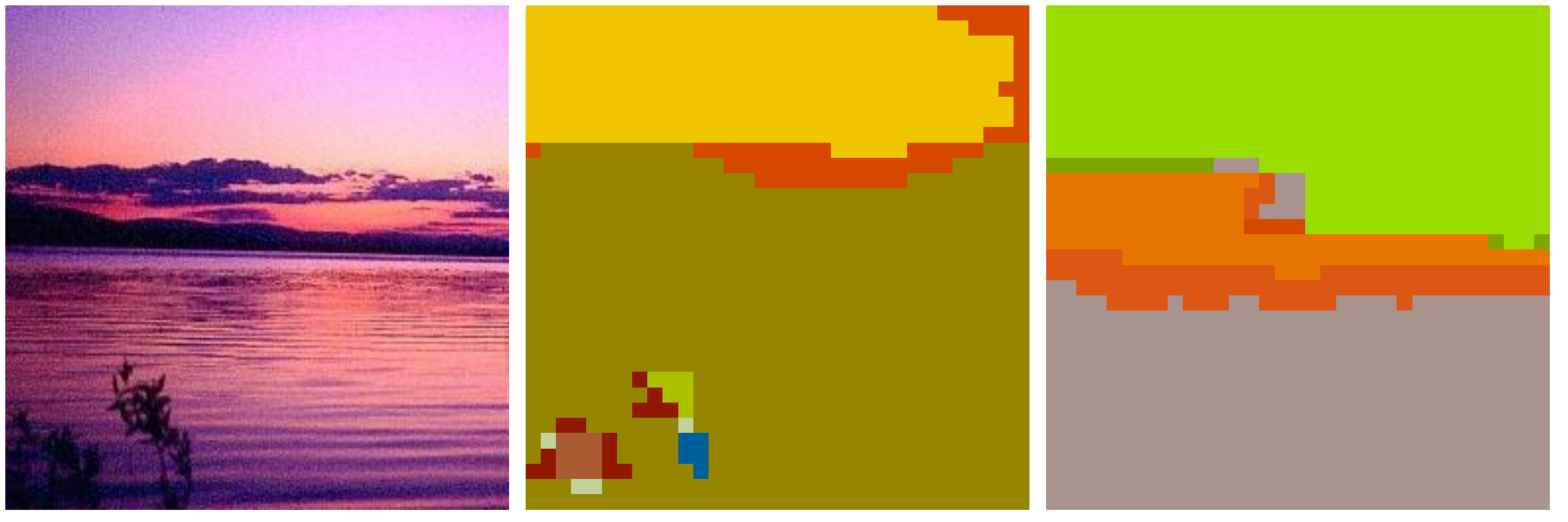}
	\caption{The features learned with a small number of time steps have stronger capability of depicting small objects compared to the features learned with a large number of time steps. \textbf{Left}: Input image. \textbf{Middle}: Output of fc7 at $t=1$. \textbf{Right}: Output of fc7 at $t=3$. The plant in the bottom-left coner of the left image is captured in the middle image but ignored in the right image.}
	\label{fig:cls123}
	%\vspace{-4mm}
\end{figure}

\subsection{Multi-Step Fusion Loss}
\label{sec:multi-step}
As shown in Figure \ref{fig:frame},
MPF-RNN is trained through a back propagation through  time (BPTT)
method which is equivalent to unfolding the recurrent network to a deep feedforward network according to the time steps by sharing parameters of weight layers across
different time steps. The conventional objective function is minimizing
the weighted sum of  cross-entropy losses between the output probability of the unfolded
network and the binary per-pixel ground truth label.

There are two disadvantages with this
objective function. First, features for small objects in an image might be inconspicuous in
the higher-level feature map due to a stack of convolution and pooling operations, which
hurts the  parsing performance since scene images have many small objects. Secondly, since the depth of an unfolded
feedforward model  with many  time steps is large, training
bottom layers in early time steps may suffer from a vanishing gradient problem.
%after many layers'error-propagation,
To
handle the above two disadvantages, we propose a new objective function with respect to the output feature maps from
multiple time steps. Formally, the objective function we propose is given as
%\begin{small}
  \begin{equation}
    \label{eq:loss}
    L = - \sum\limits_{(i,j) \in \mathbf{I}} {\mathbf{\omega}
        _{\mathbf{y}_{i,j} } \mathbf{h}_{\mathbf{y}_{i,j} } (\mathbf{w},\mathbf{O})}, % \\
    %\label{eq:sum-output}
    \mathbf{O} = \sum\limits_{t = 1}^T {\lambda _t \mathbf{X}^{(L)}(t)}.
  \end{equation}  %\end{small}

Here $T$ denotes the number of time steps, $\mathbf{O}$ denotes the combined feature of the top layer's output at each time step, and
$\mathbf{h}_{\mathbf{y}_{i,j} } (\mathbf{w},\mathbf{O})$ denotes the logarithmic prediction
probability produced by a classifier which maps from input
$\mathbf{O}$ to per-pixel category label $\mathbf{y}_{i,j}$. The specific formulation of $\mathbf{h}_{\mathbf{y}_{i,j} } (\mathbf{w},\mathbf{O})$ depends on the chosen classifier for predicting per-pixel label, such as MLP, SVM and logistic regression classifier, etc.

In our method, the
classifier contains a deconvolution layer which resizes the discriminative representation $\mathbf{O}$ to be
the same size as the input image for producing dense
predictions, followed by a softmax loss layer to produce per-pixel categorical
prediction probabilities. $\mathbf{w}$ denotes the parameters in the deconvolution
layer. Note that $\mathbf{y}_{i,j} \in \{1, \ldots, L\}$ denotes the category
of the pixel at location $(i,j)$.  $\lambda_t$ balances the importance of feature
maps at time step $t$ and $\mathbf{\omega}$ is the class weight vector. We further discuss them in the experiments.

The benefits of using Eqn. \eqref{eq:loss} are two-fold. On one hand, it improves MPF-RNN's discriminability for small objects. Because the output features at later time steps have increasingly larger receptive field (RF) which receives wider-range context information,  they may fail to distinguish small objects. In contrast, output features at earlier time steps have smaller RF and are more discriminative in capturing small objects. Therefore the combination of features in the last time step with those of ealier time steps helps the model identify small objects. On the other hand, for deep models with hundreds or thousands layers and large time steps, it may avoid the gradient vanishing problem in bottom layers at early time steps by giving shorter paths from them to the loss layer. Note that the multi-step fusion used in our method is different from the feature fusion method used in FCN~\cite{fullyconvseg}. While FCN combines
feature maps from hidden layers in a CNN, we combine the outputs at different
time steps in an RNN model, which retains stronger context modeling capabilities.

\subsection{Discussions}
\label{sec:discussion}
To better understand MPF-RNN, we compare it
with several existing RNN models. Conventional RNN is trained with a sequence of inputs $\{x^{(t)}\}_{t=1:T}$ by computing the following sequences:
%\begin{equation*}
$h^{(t)} = s(Ux^{(t)}+Wh^{(t-1)})$ and $y^{(t)}=z(Vh^{(t)})$,
% \ \ \ \
% y^{(t)}=z(Vh^{(t)})
%\end{equation*}
\\where $h^{(t)} \mbox{and}\ y^{(t)}$ are the hidden layer and output layer at time step $t$, respectively, while $U \mbox{and}\ W$ are weights matrices between the input and hidden layers, and among the hidden units themselves, respectively. $V$ is the output matrix between hidden and output layers. $s(\cdot) \mbox{ and} \ z(\cdot)$ are activation functions. Compared with the conventional RNNs, MPF-RNN has multiple recurrent connections from the output layer to hidden layers (as indicated by Eqn. (\ref{eq:mpf})) and moreover, MPF-RNN combines the outputs of multiple time steps as the final output (see Eqn.~\eqref{eq:loss}). 

Compared with those two recently proposed RNN-based models~\cite{pinheiro2013recurrent,RCNN}, which have one recurrent connection from the output layer to the input layer and layer-wise
self-recurrent connections, respectively, our
model can be deemed as a generalization by allowing more general recurrent
connections. Specifically, when $1 \in S$ (recall $S$ is the index set of selected layers with which the output layer has feedback connections), there will be a recurrent connection from the output
layer to the input layer in our model, as in~\cite{pinheiro2013recurrent}; when $L \in S$, there will be self-recurrent connections, as in \cite{RCNN}. By utilizing
context information in different layers, our model
has a stronger capability to learn discriminative features for each pixel. We also highlight the main difference of network architecture between MPF-RNN and RCN~\cite{recombinator}. RCN is still essentially a feed-forward network which utilizes features in higher layers to aggregate features in bottom layer for facial keypoint localization problem, while MPF-RNN is a novel RNN architecture for better modeling the long-range context information in scene parsing problem.

The way to build multiple recurrent connections from the top layer to  subsequent
layers is a reminiscent of fully recurrent nets (FRN)~\cite{williams1989learning} which is an MLP with each
non-input unit receiving connections from all the other units.  Different from FRN, we employ \emph{convolution}/\emph{deconvolution} layers to model the recurrent connections in MPF-RNN so as to preserve the spatial dependencies of 2D images. Besides, we do not build recurrent connections from the output
layer to every subsequent layer, since the neighboring layers contain redundant
information and the ``fully recurrent'' way is prone to over-fitting in scene
parsing tasks. In addition, we perform multi-step fusion to improve the final performance, while FRN only uses the output of the final layer. To the best of our knowledge, we are among the first to apply
multiple convolutional recurrent connections for solving scene parsing problems.

\section{Experiments}
\label{sec:experiment}
\subsection{Experiment Settings and Implementation Details}
\label{sec:misc.}
\subsubsection{Evaluation Metrics} Adopted by most previous
works as evaluation metrics, the per-pixel accuracy (PA) and the average
per-class accuracy (CA) are used. PA is defined as the percentage of all correctly
classified pixels while CA is the average of all category-wise accuracies.
\subsubsection{Baseline Models}  Following~\cite{chen2014semantic}, we use a variant of the ImageNet
pre-trained VGG16 network as the baseline model and fine-tune it on four scene parsing
datasets, including SiftFlow, Barcelona, CamVid and Stanford Background. Here, fully connected (FC) layers (fc6 and fc7) are replaced by
convolution layers.
%The last two pooling layers are skipped and the following convolution layers use ``atrous algorithm''~\cite{mallat1999wavelet} to keep convolution operations sound.
To accelerate the dense labeling prediction, the kernel size of fc6 is reduced to $3\times 3$ and the number of channels at FC layers is
reduced to 1,024. This model is referred to as ``VGG16-baseline'' in the
following experiments. Note that the VGG16-baseline model is equivalent to the MPF-RNN
model with time step equal to 1. For simplicity, in the following text, we denote conv5 as conv5\_1, conv4 as conv4\_1 and conv3 as conv3\_1.
%Also to relief the burden of heavy notation, when no confusion is incurred, we replace the layer indexes in $S$ with the corresponding layers names.

\subsubsection{Multi-Path Feedback} A practical problem with using MPF-RNN is how to choose the layer set $S$, to which the recurrent
connections from the top layer are constructed. There are two thumb rules. First, the  layers at very bottom should not be chosen since they
generally learn basic and simple patterns like edges, circles and dots, in which
global context is not helpful. Secondly, it should be avoided to choose too many
neighboring layers so that abundant information may be reduced in the features learned
from neighboring layers. Following the above two rules, we conduct experiments
with various $S$ on the validation set of SiftFlow and choose the one with best
validation performance. The comparative results are shown in Table~1, from which we can
see that choosing $S = \{\text{conv}4, \text{conv}5, \text{fc}6, \text{fc}7\}$ gives the
best performance.

\subsubsection{Multi-Step Fusion} There is a trade-off between the computation efficiency and model
complexity when choosing the number of time steps for unfolding MPF-RNN. A deeper network can learn more complex features and give better performance. On the other hand, a large model is time consuming to train and test, which turns out to be problematic when
applied to real-time scenarios, \emph{e.g.} automatic driving. Table~\ref{table:msf} compares the
performance of MPF-RNN with different time steps on SiftFlow, from which we observe that
among other time steps, $T=3$ achieves the best performance with fast speed. Therefore, we
choose $T = 3$ as the default time step for all datasets. Besides, we set $\lambda_1=\lambda_2=0.3$ and $ \lambda_3 = 1$ throughout our experiments. As an alternative, we can also see $\lambda_t$ as learnable parameters and train them jointly with the whole model end-to-end. However, we do not explore this method in the paper.

\subsubsection{Hyperparameters} The hyperparameters introduced by MPF-RNN, including $S, T$ and $\lambda$ are fine-tuned on the validation set of SiftFlow as introduced above and then fixed for other datasets where MPF-RNN uses VGG16 network. Our experiments verified that such values of hyperparameters are optimum in all datasets.
% {\color{red}In our nips submission, some reviewer want to see results of different hyperparameters on all datasets}.

\subsubsection{Loss Re-Weighting } Since in scene parsing tasks, the class distribution is
extremely unbalanced, it is common to re-weight different classes during training to attend rare classes~\cite{farabet2013learning,dag}. In our model, we adopt the reweighting strategy by~\cite{dag} because of its simplicity and effectiveness. Briefly, the weight for class
$\mathbf{y}_{i,j}$ is defined as
$\omega _{\mathbf{y}_{i,j} } = 2^{\left\lceil {\log 10(\eta /f_{\mathbf{y}_{i,j}
		} )} \right\rceil }$ where $f_{\mathbf{y}_{i,j}}$ is the frequency of class
$\mathbf{y}_{i,j}$ and $\eta$ is a dataset-dependent scalar, which is defined according to 85\%/15\% frequent/rare classes rule.

\subsubsection{Fine-Tuning Strategy} Our model is fine-tuned over target datasets
using the stochastic gradient descent algorithm with momentum. For models using VGG16 network, settings of hyper-parameters including learning
rate, weight decay and momentum follow~\cite{parsetnet}. The reported results
are based on the model trained in 40 epochs. Data augmentation is used to reduce the risk of over-fitting and improve the generalization
performance of deep neural network models. To make a fair comparison with other state-of-the-art
methods, we only adopt the common random horizontal flipping and cropping during training.

\subsubsection{Computational Efficiency} On a NVIDIA Titan X GPU, the training of MPF-RNN (the model in Table 3) on SiftFlow dataset finishes in about 6 hours and the testing time for an image with the resolution of 256$\times$256 is 0.06s.
\begin{table}
	\caption{Comparative study of effects of different recurrent connections on final
		performance of MPF-RNN over SiftFlow dataset. The time step of MPF-RNN is fixed as 2
		in all experiments in this table. The best results are shown in bold.}
	\label{table:mpf}
	\centering
	\begin{tabular}{lcc} \hline Recurrent connections       & {PA(\%)}       &{CA(\%)}  \\
		\hline
		\hline
		VGG16-baseline                                & 84.7           & 51.5  \\
		\hline
		\{fc7\}                                       & 85.4           & 55.1 \\
		\{fc6, fc7\}                                  & 85.9           & 55.5 \\
		\{conv5, fc6, fc7\}                           & 86.2           & 55.8 \\
		\{conv4, conv5, fc6, fc7\}                    & \bf{86.4}           & \bf{56.3} \\
		\{conv3, conv4, conv5, fc6, fc7\}             & 85.9           & 55.7 \\
		\hline
	\end{tabular}
\end{table}

\subsection{Results}
\label{sec:results}
We test MPF-RNN on five challenging scene parsing benchmarks, including
SiftFlow~\cite{siftflow}, Barcelona~\cite{tighe2010superparsing}, CamVid~\cite{brostow2008segmentation}, Stanford Background~\cite{gould2009decomposing} and ADE20K. We report the quantitative results here and more qualitative results of MPF-RNN are given in the Supplementary Material.

\subsubsection{SiftFlow}
\label{sec:siftflow}
The SiftFlow dataset~\cite{siftflow} consists of 2,400/200 color images with 33
semantic labels for training and testing.

\begin{table}
	\caption{Comparative study of effects of different recurrent time steps and multi-step fusion (MSF) on final performance of MPF-RNN over SiftFlow dataset. $S = \{\text{conv}4, \text{conv}5, \text{fc}6, \text{fc}7\}$ herein.}
	\label{table:msf}
	\centering
	\begin{tabular}{@{}lcccc@{}}
		\hline
		& \multicolumn{2}{c@{}}{w/ MSF} & \multicolumn{2}{c@{}}{w/o MSF} \\
		\cmidrule(lr){2-3}
		\cmidrule(lr){4-5}
		Time steps  & {PA(\%)}  &{CA(\%)}   & {PA(\%)}  &{CA(\%)}  \\
		\hline
		\hline
		VGG16-baseline                & 84.7      & 51.5      & N.A.      & N.A.  \\
		\hline
		T = 2                         & 86.4      & 56.3      & \bf{86.0} & \bf{55.5} \\
		T = 3                         & \bf{86.9} & \bf{56.5} & 85.2      & 55.1 \\
		T = 4                         & 86.8      & 55.9      & 84.7      & 54.8 \\
		\hline
	\end{tabular}
\end{table}

\emph{Model Analysis}\quad We analyze the MPF-RNN by
investigating the effects of its two important components separately,
\textit{i.e.} multi-path feedback and multi-step fusion.
Table \ref{table:mpf} lists the performance of MPF-RNN, as well
as the baseline models when different recurrent connections are used. The number of time
steps is set as 2 for all MPF-RNN models and combination weights
are $\lambda_1 = 0.3$ and
$\lambda_2 = 1$. Compared with the baseline model which has achieved
84.7\%/51.5\% PA/CA on this dataset, adding recurrent connections to
multiple subsequent layers significantly improves the performance in
terms of PA and CA. Only adding one recurrent connection to the layer
fc7 can increase the performance to 85.4\%/55.1\%, which proves
the benefit of global context modulating to the performance. Continuing adding recurrent connections to conv5 and conv4
consistently improves the performance. The reason for the continuous improvement is the
sufficient utilization of the context information in learning context-aware
features in different hidden layers. Based on above experimental
results, it is verified that multi-path feedback is beneficial
for boosting the performance of scene parsing systems. We also note that
the performance tends to slow down its increasing (although still much
higher than the baseline model) when adding too many recurrent
connections ($S = \{\text{conv4, conv5, fc6, fc7}\}$). Such a phenomenon
implies the overfitting due to the increased number of
parameters.
\begin{table}
	\caption{Comparison with the state-of-the-art methods on SiftFlow.}
	\label{table:siftflow}
	\centering
	\begin{tabular}{lcc} \hline Methods       & {PA(\%)}       &{CA(\%)}  \\
		\hline
		Tighe \emph{et al.}~\shortcite{tighe2010superparsing}                      & 79.2            & 39.2   \\
		Sharma \emph{et al.}~\shortcite{sharma2014recursive}                     & 79.6            & 33.6   \\
		Singh \emph{et al.}~\shortcite{singh2013nonparametric}                     & 79.2            & 33.8   \\
		Yang \emph{et al.}~\shortcite{yang2014context}                       & 79.8            & 48.7   \\
		Liang \emph{et al.}~\shortcite{RCNN}                      & 84.3            & 41.0   \\
		Long \emph{et al.}~\shortcite{fullyconvseg}                       & 85.2            & 51.7   \\
		Liu \emph{et al.}~\shortcite{parsetnet}                       & 86.8            & 52.0   \\
		Shuai \emph{et al.}~\shortcite{dag}                      & 85.3            & 55.7   \\
		\hline
		MPF-RNN(ours)                           & \bf{86.9}       & \bf{56.5} \\
		\hline
	\end{tabular}
\end{table}

Table \ref{table:msf} shows the effects of different time steps and multi-step
fusion on the performance. We conduct experiments when $T =1,2,3, 4$ by setting $S = \{\text{conv4, conv5, fc6, fc7}\}$. Concluded from Table~\ref{table:msf}, the PA/CA
consistently improves from $T = 1$(VGG16-baseline) to $T = 3$. Such improvement
is attributed to the more complex discriminative features learned using larger
time steps, which is proportional to the depth of the feedforward deep model
after MPF-RNN is unfolded. The improvement is also consistent with observations in \cite{vgg,googlenet,residual} that
deeper and larger models have stronger feature representation
capabilities compared with shallow ones. When $T=4$, the
performance is worse than that when $T=3$ due to the overfitting problem.

We also
conduct experiments when only the top layer's feature in the last
time step is used as the input to the classifier. Specifically, we set
$\lambda_t=0, t=2,\cdots,T$ and $\lambda_T = 1$ and keep the other conditions
unchanged. It is observed from Table~\ref{table:msf} that under any time steps, the PA/CA are worse than those
when multi-step fusion is applied. There are two reasons for the
inferior performance. Updating bottom layers parameters is insufficient since
it takes many hidden layers to propagate the error message from the output layer
to bottom layers. In addition, as illustrated in Figure \ref{fig:cls123}, the output layer at
the last time step might ignore small objects, which could otherwise be complemented by
using multi-step fusion. Above experiment demonstrates the effectiveness of
multi-step fusion. Note that although the conclusions made in model analysis are based on the results on SiftFlow, we have verified through experiments that they also hold on other three datasets where MPF-RNN uses VGG16 network.

\emph{Comparison with State-of-the-Art}\quad The comparison results of
MPF-RNN with other state-of-the-art methods are shown in Table~\ref{table:siftflow}, from which we can see that MPF-RNN achieves the best
performance against all the compared methods. Specifically, our method significantly outperforms Pinheiro \emph{et al.}~\cite{pinheiro2013recurrent} and
Liang \emph{et al.}~\cite{RCNN} by increasing the PA/CA by 9.2\%/26.7\% and 2.6\%/15.5\%,
respectively. Note that ~\cite{pinheiro2013recurrent} and ~\cite{RCNN}
are all RNN-based methods. The remarkable improvement of MPF-RNN over them is attributed to MPF-RNN's powerful
context modeling strategy by constructing multiple recurrent connections in various
forms. Compared with~\cite{parsetnet}, MPF-RNN achieves much better CA due to its superior capability of distinguishing small objects.

\subsubsection{ADE20K}
To further verify scalability of the MPF-RNN  to large-scale dataset and deeper networks, we conduct experiments using ResNet101~\cite{residual} on recently released ADE20K dataset~\cite{ade20k} which also serves as the dataset of scene parsing challenge in ILSVRC16\footnote{\scriptsize{https://image-net.org/challenges/LSVRC/2016/\#sceneseg}}. Containing 20K/2K/3K fully annotated scene-centric train/val/test images with 150 classes, ADE20K has more diverse scenes and richer annotations compared to other scene parsing dataset, which make it a challenging benchmark. Following \cite{deeplabv2}, all convolution layers in original ResNet101 \cite{residual} after conv3\textunderscore \text{4} are replaced with dilated convolution layers to compute dense scores at stride of 8 pixels, followed by a deconvolution layer to reach original image resolution. This model is used as a strong baseline and referred to as ResNet101-Baseline. Since the test set is not available till submission of the paper, we use the val set to test the performance of MPF-RNN. In order to fine-tune hyperparameters, we randomly extract 2K images from train set as our validation data and retrain our model using whole train set after fixing hyperparameters. Through validation, we set $S = \{\text{conv}4\textunderscore 1, \text{conv}4\textunderscore 9, \text{conv}4\textunderscore 17, \text{conv}5\textunderscore 1, \text{conv}5\textunderscore 3\}$ and the value of $T$ and $\lambda$ are the same as those in the last section. MPF-RNN and ResNet101-Baseline are trained for 20 epochs by fine-tuning ResNet101\footnote{\scriptsize{$\textrm{https://}\textrm{github}.\textrm{com/}\textrm{KaimingHe/}\textrm{deep-residual-networks}$}} on ADE20K with the same solver configurations of \cite{deeplabv2}. Data augmentation \emph{only} include random horizontal flipping and cropping. We do not use loss re-weighting in this dataset.
\begin{small}
\begin{table}
	\caption{Comparison results of MPF-RNN on ADE20K val set. The results of FCN, SegNet and DilatedNet are referred from the reported number in~\protect\cite{ade20k} }
	\label{table:ade20k}
	\centering
	\begin{tabular}{lcc} \hline Methods       & {PA(\%)}       &{mIOU(\%)}  \\
		\hline
		FCN~\shortcite{fullyconvseg}   & 71.32          & 29.39 \\
		SegNet~\shortcite{segnet}                    & 71.0            & 21.64 \\
		DilatedNet~\shortcite{dilatenet}                & 73.55            & 32.31  \\
		Res101-Baseline        &  73.71           & 32.65   \\
		\hline
		MPF-RNN \small{(ours)}    & \bf{76.49}        & \bf{34.63} \\
		\hline
	\end{tabular}
\end{table}
\end{small}

In Table \ref{table:ade20k}, we compare the performance of MPF-RNN and baseline models. Following the evaluation metric in ILSVRC16 scene parsing challenge, we compare the PA and mean IOU (the mean value of intersection-over-union between
the predicted and ground-truth pixels across all classes) between different models. It is observed from Table \ref{table:ade20k} that Res101-Baseline has achieved better performance compared to other models based on VGG16 due to its superior deep architecture. By using MPF-RNN, MPF-RNN significantly surpasses this strong baseline model by 2.78\%/1.98 in terms of PA/mIOU, again demonstrating the capability of MPF-RNN to improve the scene parsing performance of deep networks.

\subsubsection{Results on Other Datasets}
We further evaluate MPF-RNN on  Barcelona, Stanford Background and Camvid datasets. MPF-RNN achieves the best performance on these three datasets: (i) Stanford Background: $86.6\%$ (PA) and $79\%$ (CA), outperforming state-of-the-art with $3\%$ (PA) and $5\%$ (CA) absolutely; (ii) Barcelona: $78.5\%$ (PA) and $29.3\%$ (CA) higher than state-of-the-art with around $4\%$ (PA) and $5\%$ (CA) absolutely; (iii) Camvid: $92.8\%$ (PA) and $82.3\%$ (CA), outperforming state-of-the-art with $1\%$ (PA) and $4\%$ (CA) absolutely.
Due to space limits, more numbers and details of experimental results on these three datasets are deferred to the supplementary material.
%The comparison between MPF-RNN and state-of-the-art methods are listed in table~\ref{table:barcelona}, table~\ref{table:stanford} and table~\ref{table:camvid}, respectively.
%\vspace{-3mm}
\section{Conclusion}
\label{sec:conclusion}
We proposed a novel \textbf{M}ulti-\textbf{P}ath \textbf{F}eedback recurrent neural network (MPF-RNN) for better solving scene
parsing problems. In MPF-RNN, multiple recurrent connections are constructed
from the output layer to hidden layers. To  learn features discriminative for small objects, output features at
each time step are combined for per-pixel prediction in MPF-RNN. Experimental results over
five challenging scene-parsing datasets, including SiftFlow, ADE20K, Barcelona, Stanford Background,
and CamVid clearly demonstrated the advantages of MPF-RNN for scene parsing.

\paragraph{Acknowledgement} Xiaojie Jin is immensely grateful to Bing Shuai at NTU and Wei Liu at UNC - Chapel Hill for valuable discussions. The work of Jiashi Feng was partially supported by National University of Singapore startup grant R-263-000-C08-133 and Ministry of Education of Singapore AcRF Tier One grant R-263-000-C21-112.
{
\small
\bibliographystyle{aaai}
\bibliography{mybibfile}
}
\newpage


\section*{Supplementary material}

# Multi-Path Feedback Recurrent Neural Network for Scene Parsing
## – Supplymentary Material

## Introduction

In this supplymentary material, we present the qualitative scene parsing results on the validation set of SiftFlow, and introduce in detail the experimental results on three datasets: Barcelona, Camvid and Stanford Background.

## Example Scene Parsing Results

Please refer to Figure 1.

## Experimental Results

### Barcelona

The Barcelona dataset has 14871/279 training/testing color images with 170 semantic categories. Following (Farabet et al. 2013), we conduct experiments either with loss re-weighting (LRW) or not on this dataset. Results are reported and compared with other methods in Table 1. When LRW is used, we observe that MPF-RNN achieves the best CA among all methods but with unsatisfied PA. PA's decreasing can be explained by the reason that enhanced capability of discriminating small objects undermines the overall pixelwise classification accuracy, which is justified from the perspective of object recognition. Note the decreasing of PA when using LRW is also observed in (Farabet et al. 2013), which is outperformed by our methods with a large margin as shown in Table 1. When LRW is not used, *i.e.* setting the loss weights for all classes to be 1, the PA is greatly improved to be the best among other methods. Although the CA of MPF-RNN is not as good as (Shuai et al. 2015) which uses the LRW, MPF-RNN still performed much better than the rest methods. Therefore, we expect to further improve the CA of MPF-RNN by using a better loss re-weighting methods in our future work.

### Stanford Background

The Stanford Background dataset contains 715 color images labeled with 8 semantic categories. Most of images have a resolution of $320 \times 240$ pixels. Following (Liang, Hu, and Zhang 2015), we report the performance of MPF-RNN using 5-fold cross validation, each of which has 572/143 training/testing images. A summary of best methods on this

Table 1: Comparison with the state-of-the-art methods on Barcelona dataset. Results of MPF-RNN either with loss re-weighting (LRW) or not are reported

| Methods | PA(%) | CA(%) |
| --- | --- | --- |
| Farabet *et al.* (2013) (w/ LRW) | 46.4 | 12.5 |
| Farabet *et al.* (2013) (w/o LRW) | 67.8 | 9.5 |
| Tighe *et al.* (2010) | 66.9 | 7.6 |
| Shuai *et al.* (2015) | 74.6 | 24.6 |
| MPF-RNN (w/ LRW) | 60.1 | **29.3** |
| MPF-RNN (w/o LRW) | **78.5** | 19.5 |

Table 2: Comparison with the state-of-the-art methods on Stanford Background dataset.

| Methods | PA(%) | CA(%) |
| --- | --- | --- |
| Gould *et al.* (2009) | 76.4 | N.A. |
| Tighe *et al.* (2010) | 77.5 | N.A. |
| Socher *et al.* (2011) | 78.1 | N.A. |
| Eigen *et al.* (2012) | 75.3 | 66.5 |
| Singh *et al.* (2013) | 74.1 | 62.2 |
| Lempitsky *et al.* (2011) | 81.9 | 72.4 |
| Farabet *et al.* (2013) | 81.4 | 76.0 |
| Pinheiro *et al.* (2013) | 80.2 | 69.9 |
| Sharma *et al.* (2014) | 81.9 | 73.6 |
| Mostajabi *et al.* (2015) | 82.1 | 77.3 |
| Liang *et al.* (2015) | 83.1 | 74.8 |
| MPF-RNN(ours) | **86.6** | **79.6** |

dataset is provided in Table 2, from which one can again observe that MPF-RNN performs better than other methods with a significant margin. Particularly, MPF-RNN continues outperforming (Liang, Hu, and Zhang 2015) which has only self-recurrent connections by 3.5%/4.8% in terms of PA/CA, demonstrating the efficacy of MPF-RNN.

### Camvid

The CamVid dataset contains 701 color images from 32 semantic classes extracted from driving videos at daytime and dusk. The resolution of images is $960 \times 720$, which is much larger than the other two datasets. Following the standard training-test split protocol, there are 468/233 images used for training and testing respectively. Following

Table 3: Comparison with the state-of-the-art methods on CamVid dataset.

| Methods | PA(%) | CA(%) |
|---|---|---|
| Tighe *et al.* (2010) | 78.6 | 43.8 |
| Sturgess *et al.* (2009) | 83.8 | 59.3 |
| zhang *et al.* (2012) | 82.1 | 55.4 |
| Bulo *et al.* (2014) | 82.1 | 56.1 |
| Ladicky *et al.* (2010) | 83.8 | 62.5 |
| Tighe *et al.* (2010) | 83.9 | 62.5 |
| Shuai *et al.* (2015) | 91.6 | 78.1 |
| MPF-RNN(ours) | **92.8** | **82.3** |

previous works (Shuai et al. 2015; Tighe and Lazebnik 2010; Brostow et al. 2008), we only train and test MPF-RNN on the most common 11 categories. Table 3 summarizes the performance of MPF-RNN and other methods. MPF-RNN achieves the best performance against all the other methods. Notably, over the previous best method on this dataset, MPF-RNN improves the PA/CA by 1.2%/4.2%.

Figure 1: Qualitative results on the validation set of SiftFlow dataset. Those four images in each column correspond to input image, the score map output by baseline CNN model, the score map output by MPF-RNN and the ground-truth pixel-wise label in the order from top to bottom, respectively. Best viewed in color.


\end{document}